\documentclass[conference]{IEEEtran}
\IEEEoverridecommandlockouts
\usepackage{amsmath,amssymb,amsfonts}
\usepackage{algorithmic}
\usepackage{graphicx}
\graphicspath{{images/}}
\usepackage{xcolor}
\def\BibTeX{{\rm B\kern-.05em{\sc i\kern-.025em b}\kern-.08em
    T\kern-.1667em\lower.7ex\hbox{E}\kern-.125emX}}
\begin{document}

\title{VT-ADL: A Vision Transformer Network for Image Anomaly Detection and Localization\thanks{This work is partially supported by beanTech srl.}}
\author{\IEEEauthorblockN{Pankaj Mishra}
\IEEEauthorblockA{University of Udine, Italy\\
Email: mishra.pankaj@spes.uniud.it}
\and
\IEEEauthorblockN{Riccardo Verk}
\IEEEauthorblockA{University of Udine, Italy\\
Email: verk.riccardo@spes.uniud.it}
\and
\IEEEauthorblockN{Daniele Fornasier}
\IEEEauthorblockA{beanTech srl, Italy\\
	Email: daniele.fornasier@beantech.it}
\and
\IEEEauthorblockN{Claudio Piciarelli}
\IEEEauthorblockA{University of Udine, Italy\\
Email: claudio.piciarelli@uniud.it}
\and
\IEEEauthorblockN{Gian Luca Foresti}
\IEEEauthorblockA{University of Udine, Italy\\
Email: gianluca.foresti@uniud.it}
}

\maketitle

\begin{abstract}
We present a transformer-based image anomaly detection and localization network. Our proposed model is a combination of a reconstruction-based approach and patch embedding. The use of transformer networks helps preserving the spatial information of the embedded patches, which is later processed by a Gaussian mixture density network to localize the anomalous areas.
In addition, we also publish BTAD, a real-world industrial anomaly dataset. Our results are compared with other state-of-the-art algorithms using publicly available datasets like MNIST and MVTec.
\end{abstract}

\begin{IEEEkeywords}
Anomaly Detection, Anomaly segmentation, Vision transformer, Gaussian density approximation, Anomaly dataset
\end{IEEEkeywords}

\section{Introduction}


In computer vision, an anomaly is any image or image portion which exhibits significant variation from the pre-defined characteristics of normality. Anomaly Detection is thus the task of identifying these novel samples in supervised or unsupervised ways. A system which can perform this task in an intelligent way is hugely in demand, as its applications range from video surveillance~\cite{Piciarelli2008-trajectory} to defect segmentation~\cite{piciarelli2018vision,chen15-distributed}, inspection~\cite{piciarelli2018vision}, quality control~\cite{napoletano2018anomaly}, medical imagining~\cite{Xingjun-etal}, financial transactions~\cite{yu2020stock} etc. As it can be seen from the examples, anomaly detection is particularly significant in the industrial field, where it can be used to automatically identify defective products.

Recent efforts have been made to improve the anomaly detection task in the field of deep learning. Most of the works try to learn the manifold of a single class representing normal data\cite{wulsin10-semi}, using an encoding-decoding scheme, and their output is a classification of the input image as either normal or anomaly, while fewer works deal with the task to segment the local anomalous region in an image\cite{bergmann2020uninformed}. Majorly, the methods either use a reconstruction-based approach, or learn the distribution of the latent features extracted by a pre-trained network or trained in end-to-end fashion. 

\begin{figure}[ht]
	\centering
	\includegraphics[width=0.48\textwidth]{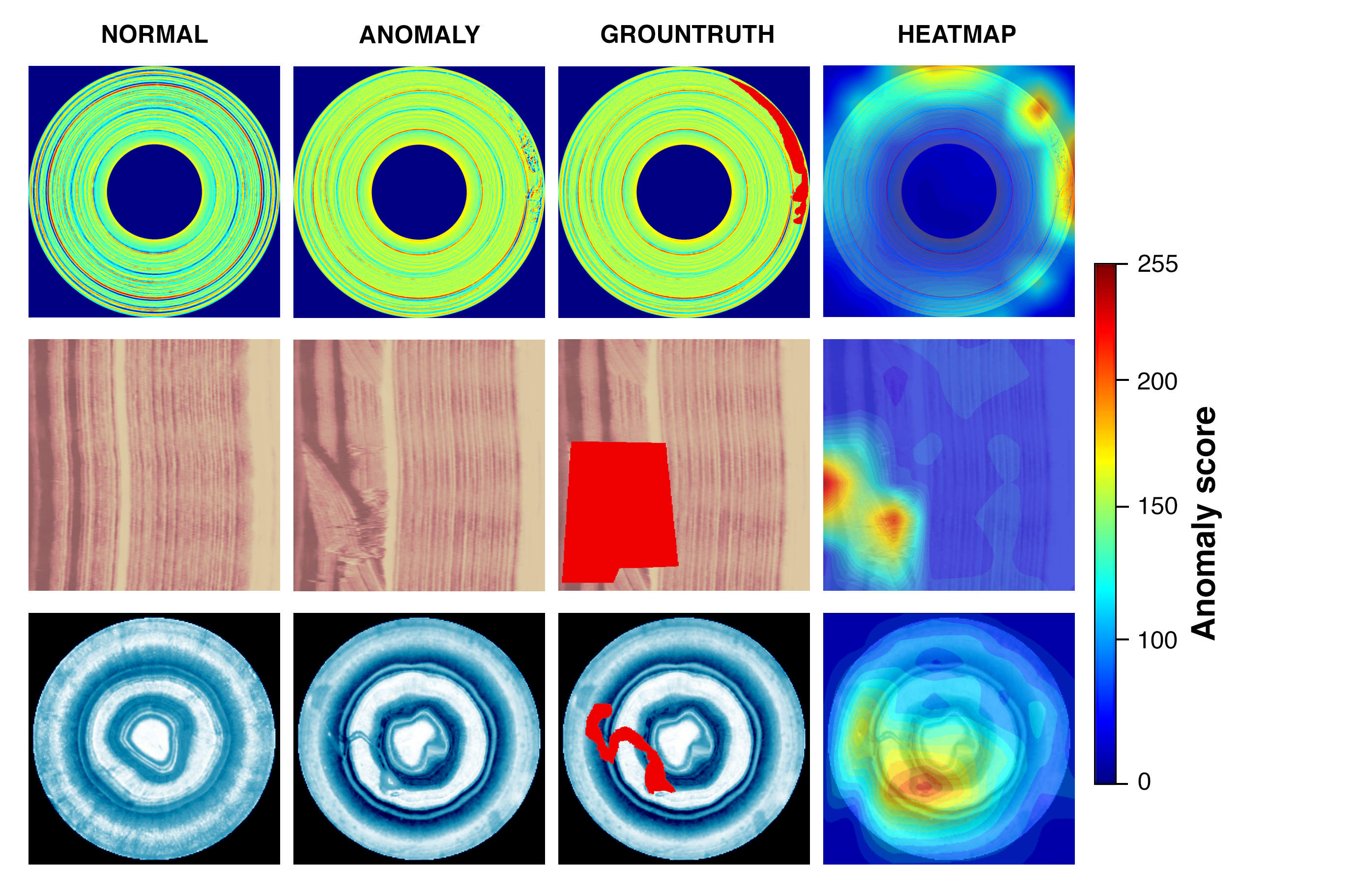}
	\caption{The three products of BTAD dataset. First column shows an example of normal images, second column shows anomalous images, third column shows the anomalous image with pixel-level ground truth labels, fourth column shows the predicted heat map by our proposed method.}
	\label{fig:bt_dataset}
\end{figure}

Motivated from the above facts and industrial needs, we developed a Vision-Transformer-based image anomaly detection and localization network (VT-ADL), which learns the manifold of normal class data in a semi-supervised way, thus requiring only normal data in the training process. 
The vision transformer network model, recently proposed by Dosovitskiy et al.~\cite{dosovitskiy2020image}, is a network designed to work on image patches trying to preserve their positional information. 
In our work we show how an adapted transformer network can be used for anomaly localization using Gaussian Approximation~\cite{bishop1994mixture, ueda1998split} of the latent features and also how different configurations can be tweaked to win some of the shortcomings of the vision transformer network.
In addition to this, we are also publishing a real-world industrial dataset (the beanTech Anomaly Detection dataset --- BTAD) for the anomaly detection task. The dataset contains a total of 2830 real-world images of 3 industrial products showcasing body and surface defects. 


\section{Related Work}

Image-based anomaly detection is not a new topic in the industrial use cases, as it has been used in many inspection and quality control schemes, however is still under investigation with modern deep learning techniques. 
Historically, several classical image processing and machine learning methods have been used to perform anomaly detection tasks, such as Bayesian networks, rule-based systems, clustering algorithms etc. However, in the recent years the trend has been shifted to the use of deep learning methods, as the convolutional layers have revolutionized this field. Most of the proposed approaches are based on image reconstruction: in this case, the network is trained to reconstruct the input image. If the network is trained on normal data only, it is assumed it will fail at properly reconstructing anomalies. Network architectures mostly consist of various configuration of autoencoders~\cite{mishra2020neural,liu2020towards,mishra2020image, Goodfellow-et-al-2016,baldi2012autoencoders} or Generative Adversarial Network (GAN)~\cite{sabokrou2018adversarially,pidhorskyi2018generative}. At image level, the simplest way is to train using MSE loss, and in turn expect higher reconstruction loss for the anomalous images. Additional information of the latent space~\cite{abati2019latent} are also used for better classification. Yet for the anomaly localization, pixel-wise reconstructed error is taken as the anomaly score. Some methods also tried to use visual attention maps~\cite{venkataramanan2020attention, liu2020towards} from the latent space. Reconstruction based methods are very intuitive and explainable, but their performance is limited when it comes to capture small localized anomalies\cite{perera2019ocgan}.


Regarding the learning method, few works adopt a fully supervised approach. It consist in training a binary classifier, in which two classes represent the normal and the anomalous data. However real-world anomaly datasets are extremely imbalanced, since the number of anomalies is typically orders of magnitude smaller than the number of normal data. This requires specialized approaches to handle data imbalance~\cite{piciarelli2019image,perera2019learning}. The majority of the solutions however rely on a semi-supervised approach, in which only normal data are available in the training step. In this case the system tries to learn a ``normality'' model from the training data and thus subsequently classify new samples as anomaly if they don't fit the model~\cite{Piciarelli2008-trajectory,baldi2012autoencoders,mishra2020neural, bergmann2019mvtec}. Recently P.Bergmann et.al~\cite{bergmann2020uninformed} developed a novel network and training scheme for both image anomaly detection and localization. The approach uses a student-teacher learning scheme and knowledge distillation for achieving state-of-the-art results and a single network for both classification and pixel level anomaly localization. The work achieves good results, but it uses a complex training scheme with high number of student networks, which again demands high computing resources for the industrial applications.
Finally, some models are based on unsupervised learning: in this case the most common approach is to use the deep network only for feature extraction and then later use some clustering algorithm, such as one-class SVM or SVDD for the final classification. Some of the works handled these two steps independently~\cite{aytekin2018clustering, erfani2016high}, while others achieved better results by doing the two steps collectively, in order to extract the best features for subsequent anomaly detection~\cite{ghafoori2020deep, chalapathy2018anomaly}.

\section{Proposed Model}

\begin{figure*}[ht]
	\centering
	\includegraphics[width=0.9\textwidth]{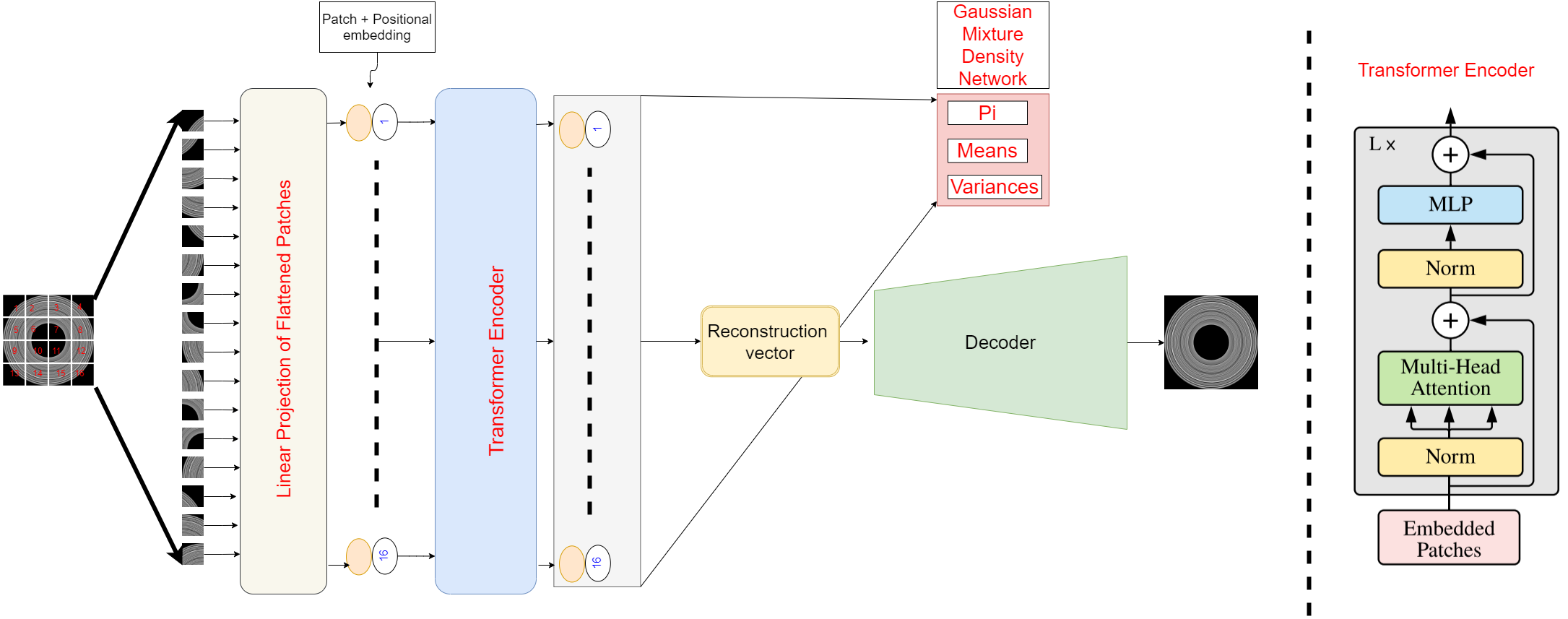}
	\caption{Left image: model overview. Image is split into patches, which are augmented with positional embedding. The resulting sequence is fed to the Transformer encoder. Then encoded features are summed into a reconstruction vector which is fed to decoder. The transformer encoded features are also fed into an Gaussian approximation network \cite{bishop1994mixture}, which is later used to localize the anomaly. Right image: detailed structure of the tranformer encoder (image from~\cite{dosovitskiy2020image}).}
	\label{fig:network}
\end{figure*}

The proposed model combines the traditional reconstruction-based methods with the benefits of a patch-based approach. The input image is subdivided in patches and encoded using a Vision Transformer. The resulting features are then fed into a decoder to reconstruct the original image, thus forcing the network to learn features that are representative of the aspect of normal images (the only data on which the network is trained). At the same time, a Gaussian mixture density network models the distribution of the transformer-encoded features in order to estimate the distribution of the normal data in this latent space. Detecting anomalies with this model automatically allows their localization, since transformer-encoded features are associated to positional information.

An overview of the model is depicted in Figure \ref{fig:network}. To handle a 2D image $X \in \mathbb{R}^{H\times W\times C}$, we break the image into a sequence of 2D patches $X_{p} \in \mathbb{R}^{N \times (P \times P \times C)}$, where $(H,W)$ is the original image resolution, $C$ is the number of channels, $(P,P)$ is the patch dimensions and $N$ is the resulting number of patches $ N = HW/P^{2}$. These patches are then embedded to a $D$-dimensional embedding space through a linear layer. Positional embedding is added to the patch embedding to preserve the positional information.
\begin{itemize}
    \item \textbf{Transformer Encoder}: The transformer encoder layer is based on the work by Vaswani et al \cite{vaswani2017attention} and its application to images by Dosovitskiy et al~\cite{dosovitskiy2020image}. 
    The input patches are first mapped to the embedding space and are augmented with positional information (eq.~\ref{eq:embedding}), then passed to a Multi-headed Self-Attention block (eq~\ref{msa_ln}) and a MLP block (eq.~\ref{mlp_ln}). Layer normalization (LN) is applied before the two blocks and residual connections are added after the two blocks. We didn't use the dropout layer throughout the network, as this causes instability in the Gaussian approximation network. MLP contains two linear layers with a GELU activation function.
    \begin{equation}
		\label{eq:embedding}
        Z_{0} = [X_{p}^{1}\textbf{E};X_{p}^{2}\textbf{E};...;X_{p}^{N}\textbf{E}] + \textbf{E}_{pos},
    \end{equation}
    $\textbf{E} \in \mathbb{R}^{(P^{2}.C)\times D},$
    $\textbf{E}_{pos} \in \mathbb{R}^{(N +1) \times D}$

    \begin{equation}
    \label{msa_ln}
    Z^{'}_{l} = MSA(LN(Z_{l-1})) + Z_{l-1}, l = 1..L
    \end{equation}

    \begin{equation}
    \label{mlp_ln}
    Z_{l} = MLP(LN(Z^{'}_{l})) + Z^{'}_{l}, l = 1..L
    \end{equation}

The final encoded patches are reshaped and projected in to a reconstruction vector via learned projection matrix. 

    
    \item \textbf{Decoder}: The decoder is used to decode the reconstruction vector back to the original image shape. It maps $\mathbb{R}^{512}\longrightarrow \mathbb{R}^{H\times W \times C}$. In our experiments with the MVTec and BTAD dataset, we used 5 transposed convolutional layers, with batch normalization and ReLU in-between, except for the last layer, we use tanh as the final non-linearity. 
    
    \item \textbf{Gaussian Mixture Density Network}: This kind of network estimates the conditional distribution $p(y|x)$ \cite{bishop1994mixture} of a mixture density model. In particular, the parameters of the unconditional mixture distribution $p(y)$ are estimated by the neural network, which takes the image embedding (conditional variable $x$) as the input. For our purpose we employ the Gaussian Mixture Model (GMM) with full co-variance matrix $\Sigma_{k}$ as the density model. The density estimate $\hat{p}(y|x)$ follows the weighted sum of $K$ Gaussian functions.
    \begin{equation}
        \hat{p}(y|x) = \sum^{K}_{k =1} w_{k}(x;\theta)\mathcal{N}(y|\mu_{k}(x;\theta), \sigma^{2}_{k}(x;\theta))
    \end{equation}
    wherein, $w_{k}(x;\theta)$ denotes the weight, $\mu_{k}(x;\theta)$ the mean, $\sigma^{2}_{k}(x;\theta)$ the variance of the k-th Gaussian. All the GMM parameters are estimated using the neural network with parameters $\theta$ and input $x$. The mixing weights of the Gaussians must satisfy the constraints $\sum^{K}_{k=1}w_{k}(x;\theta) = 1$ and $w_{k}(x;\theta) \geq$ 0 $\forall k$. This is achieved using the softmax function to the output of weight estimation:
    \begin{equation}
        w_{k}(x) = \frac{exp(a^{w}_{k}(x))}{\sum^{K}_{k=1}exp(a^{w}_{i}(x))}
    \end{equation}
    wherein $a_{k}^{w}(x) \in \mathbb{R}$ is the logit scores emitted by the neural network. Additionally, standard deviation $\sigma_{k}(x)$ must be positive. To satisfy this, a softplus non-linearity is applied to the output of the neural network.
    \begin{equation}
        \sigma_{k}(x) = log(1+exp(\beta \times x)) ; \beta =1
    \end{equation}
    Since, mean $\mu_{k}(x;\theta)$ doesn't have any constraint, we used linear layer without any non-linearity for the respective output neurons.
\end{itemize}



\section{Objective and Losses}

Training the network has two objectives: on one side we want the decoder output to resemble the network input, as in reconstruction-based anomaly detection. This forces the encoder to catch features that are relevant to describe the normal data. On the other side, the goal is to train the Gaussian mixture density network to model the manifold where the encoded features of normal images reside. For the reconstruction-based part we adopted a combination of two losses:

\begin{itemize}
    \item \textit{Mean Squared Error (MSE)}: it is a pixel-level loss, which assumes independence between pixels. MSE loss is computed as the average of the squared pixel-wise differences of the two images, and can be formally defined in terms of the Frobenius norm as  $\frac{1}{WH}\| X - \hat{X}\|_{F}^2$, where $X$ is the input and $\hat{X}$ is the output of the decoder network (respectively the original and the reconstructed image), and $W,H$ are the image width and height respectively.
    
    \item \textit{Structural Similarity Index} - The Structural Similarity Index (SSIM)~\cite{bergmannimproving} is used to measure the image similarity by considering visual properties that are lost in the standard MSE approach:
    \begin{equation}
		SSIM(x,y) = \frac{(2\mu_x\mu_y + c_1)(2\sigma_{xy} + c_2)}{(\mu_x^2 + \mu_y^2 + c_1)(\sigma_x^2 + \sigma_y^2 + c_2)}
	\end{equation}
	where, $\mu_x$, $\mu_y$, are the average values of input and reconstruction image, $\sigma_x^2$, $\sigma_y^2$ are the variance of input and reconstructed image, $\sigma_{xy}$ is their co-variance  and $c_1 , c_2$ are the two constants used for numerical stability.
\end{itemize}
    
For the Gaussian mixture density network training we used the Log-Likelihood Loss (LL). The parameter $\theta$ of the Gaussian estimation network are fitted through maximum likelihood estimation. We minimize the negative conditional log-likelihood of the normal class training data.
    \begin{equation}
        \theta^{*} = - \arg \min_{\theta} \sum^{K}_{k=1} \log p_{\theta}(y_{n}|x_{n})
    \end{equation}
    For the purpose of regularization, we also add Gaussian noise $\mathcal{N}(0,0.2)$ to the transformer embedded features before feeding it to the GMM model. Adding noise during training is seen as a form of data augmentation and regularization that biases towards smooth functions \cite{bishop1994mixture,6796981}.

Hence, the final objective function to minimize is the weighted addition of the above three losses.
\begin{equation}
\normalsize
    L(X) = -LL + \lambda_{1} MSE(X,\hat{X}) +\lambda_{2}  SSIM(X,\hat{X})
\end{equation}
wherein, $\lambda_{1} = 5$ and $\lambda_{2} = 0.5$ for all the datasets used in this study.

\section{Experimental Results}
In this section, we present the experimental results obtained by our proposed network VT-ADL. We first describe the used datasets, training and testing procedures and comparative results. We also introduce the beanTech Anomaly Detection Dataset\footnote{http://avires.dimi.uniud.it/papers/btad/btad.zip} (BTAD), a novel dataset of real-world, industry-oriented images composed of both normal and defective products. The defective images have been pixel-wise manually labeled with a ground-truth anomaly localization mask.

\begin{figure}[tb]
	\centering
	\includegraphics[width=0.5\textwidth]{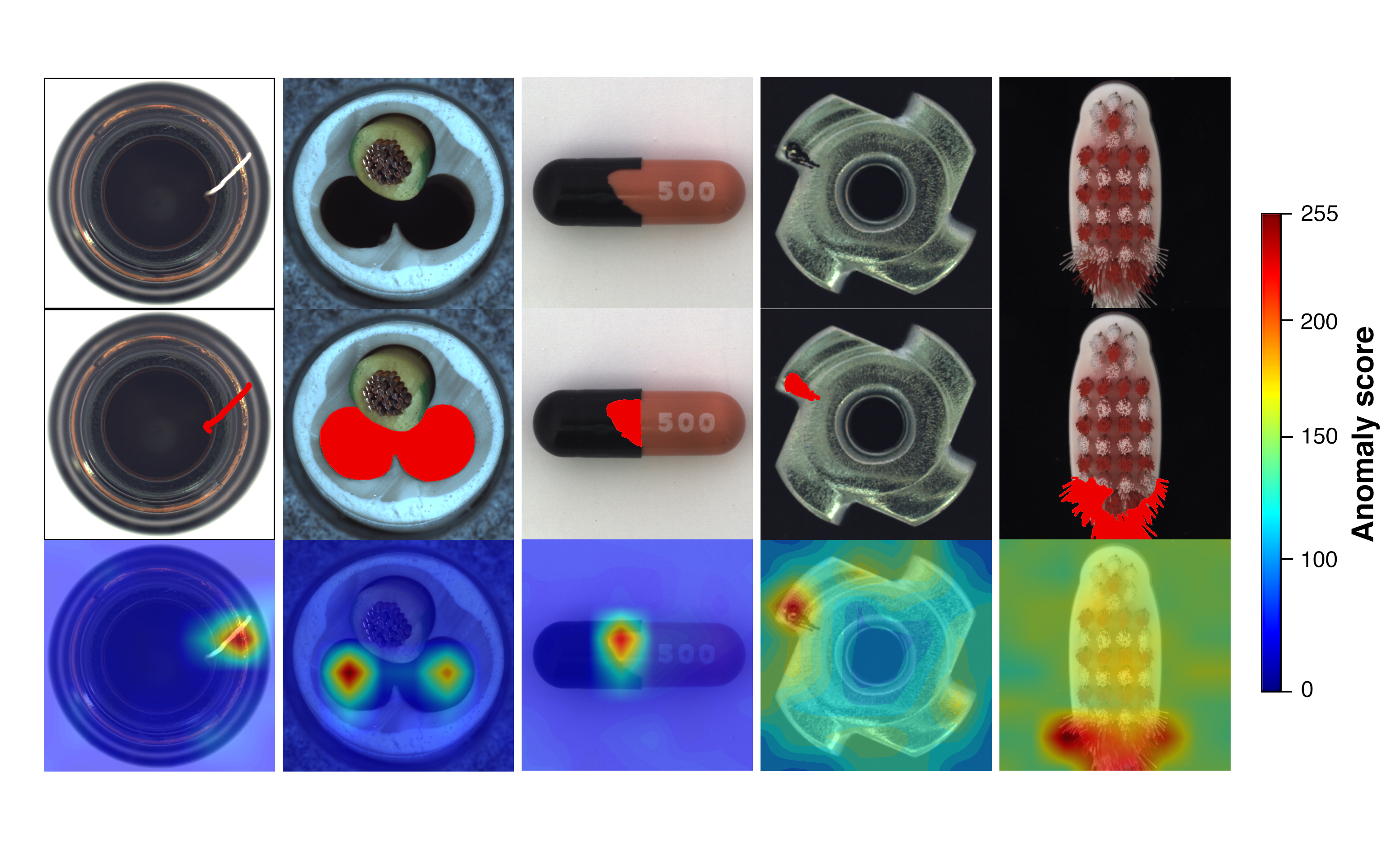}
	\caption{Anomaly detection on MVTec dataset. First row shows the actual anomalous image of bottle, cable, capsule, metal nut and brush. second row shows the actual ground truth and third row shows the generated anomaly score and anomaly localization by our method}
	\label{fig:mvtec-heatmap}
\end{figure}

\subsection{Datasets}
\begin{itemize}
	\item \textbf{MNIST}: MNIST dataset consists of 60K gray images of hand written digits. Although this dataset  was not originally developed for anomaly detection tasks, it has often been used as a baseline dataset, thus we used it to compare with other state-of-the-art approaches. For training, one class has been considered as normal, while all others as anomaly.
	
    \item \textbf{MVTec Dataset}: It's a real-world anomaly detection dataset. It contains 5,354 high-resolution color images of different textures and objects categories. It has normal and anomalous images which showcase 70 different types of anomalies of different real-world products. It contains gray scale images as well as RGB images. Gray scale images are quite common in industrial scenarios. With this dataset, all the images were first scales to $550 \times 550$ pixels and then center cropped to $512 \times 512$pixels before being passed to the model.
        
    \item \textbf{BTAD Dataset}: It contains RGB images of three industrial products. Product 1 is of $1600\times1600$ pixels, product 2 is of  $600\times600$ and product 3 is of $800\times600$ pixels in size. Product 1, 2 and 3 have 400, 1000 and 399 train images respectively. While training all the images were first scaled to 512 before passing to the model. For each anomalous image, a pixel-wise ground truth mask is given.

\end{itemize}
While training, we fed our model using the normal class data only. While testing, a combination of reconstruction losses and the maximum of the log-likelihood loss are used to perform global anomaly detection, while the log-likelihood loss alone is used for anomaly localization. In this second case, we stored the log-likelihood loss for all the patch positions and then upsample it using 2D bilinear-upsampling, to input image size, to obtain the heatmap. Then we employed the PRO (Per Region Overlap) \cite{bergmann2020uninformed,bergmann2019mvtec} as the evaluation metric for the MVTec and BTAD datasets. For computing PRO, heatmaps are first thresholded at a given anomaly score to make the binary decision for each pixel. Then the percentage of overlap with the ground truth (GT) is computed. We followed the same approach as in \cite{bergmann2020uninformed}, to find the PRO value for a large number of increasing thresholds until an average per-pixel positive rate of 30\% is reached. For the MNIST dataset, we adopted AUC (area under ROC curve) as a performance metric in order to show comparative results.

The hyper-parameters used in the training are show in table~\ref{tab:hperparameters}.
\begin{table}[h]
\centering
\begin{tabular}{|l|l|}
\hline
Adams lr rate & 0.0001 \\ \hline
Weight decay & 0.0001 \\ \hline
Batch Size & 8 \\ \hline
Epochs & 400 \\ \hline
No. of Gaussian's & 150 \\ \hline
Patch Dimension & P = 64 \\ \hline
\end{tabular}
\caption{Training hyperparameters}
\label{tab:hperparameters}
\end{table}

\subsection{Results}
Before considering the problem of anomaly localization, we tested our model on the MNIST dataset, which has been widely used as a reference dataset for anomaly detection. In this case, one class is selected as normal and anything else is considered anomalous. The anomalies are thus defined at a global level, rather than being localized in specific, possibly small image patches as in the more challenging MVTec and BTAD datasets. For this reason, anomaly detection is performed only by using the global reconstruction losses, without measuring the localization output. The results are reported in table \ref{tab:MNIST-result}, where they are compared with the performances of other popular anomaly detection techniques (results taken from \cite{abati2019latent,mishra2020image}). As it can be seen, the proposed method almost always outperforms the competitors.

\begin{table*}[h]
\centering
\begin{tabular}{|l|l|l|l|l|l|l|l|l|l|}
\hline
\textbf{Class} & \textbf{\begin{tabular}[c]{@{}l@{}}OC\\ SVM\end{tabular}} & \textbf{KDE} & \textbf{DAE} & \textbf{VAE} & \textbf{\begin{tabular}[c]{@{}l@{}}Pix\\ CNN\\ GAN\end{tabular}} & \textbf{LSA} & \textbf{\begin{tabular}[c]{@{}l@{}}Deep\\ SVDD\end{tabular}} & \textbf{\begin{tabular}[c]{@{}l@{}}Pyr.\\ AE\end{tabular}} & \textbf{VT-ADL} \\ \hline
\textbf{0} & 0.988 & 0.885 & 0.991 & 0.994 & 0.531 & 0.993 & 0.98 & 0.995 & \textbf{0.99} \\ \hline
\textbf{1} & 0.999 & 0.996 & 0.999 & 0.999 & 0.995 & 0.999 & 0.997 & 0.999 & \textbf{1} \\ \hline
\textbf{2} & 0.902 & 0.71 & 0.89 & 0.96 & 0.478 & 0.959 & 0.917 & 0.941 & \textbf{0.976} \\ \hline
\textbf{3} & 0.950 & 0.693 & 0.935 & 0.947 & 0.517 & 0.966 & 0.919 & 0.966 & \textbf{0.976} \\ \hline
\textbf{4} & 0.955 & 0.844 & 0.921 & 0.965 & 0.739 & 0.956 & 0.949 & 0.960 & \textbf{0.984} \\ \hline
\textbf{5} & 0.968 & 0.776 & 0.937 & 0.963 & 0.542 & 0.964 & 0.885 & \textbf{0.972} & 0.971 \\ \hline
\textbf{6} & 0.978 & 0.861 & 0.981 & 0.995 & 0.592 & 0.994 & 0.983 & 0.993 & \textbf{0.995} \\ \hline
\textbf{7} & 0.965 & 0.884 & 0.964 & 0.974 & 0.789 & 0.980 & 0.946 & \textbf{0.993} & 0.99 \\ \hline
\textbf{8} & 0.853 & 0.669 & 0.841 & 0.905 & 0.340 & 0.953 & 0.939 & 0.895 & \textbf{0.974} \\ \hline
\textbf{9} & 0.995 & 0.825 & 0.96 & 0.978 & 0.662 & 0.981 & 0.965 & 0.989 & \textbf{0.99} \\ \hline
\textit{Mean} & \textit{0.95} & 0.81 & 0.94 & 0.97 & 0.62 & 0.97 & 0.948 & 0.97 & \textbf{0.984} \\ \hline
\end{tabular}
\caption{AUC results of anomaly classification using MNIST, Each row shows the normal class of the trained model. Comparative results are taken from \cite{abati2019latent,mishra2020image}}
\label{tab:MNIST-result}
\end{table*}

\begin{table*}[h]
\centering
\begin{tabular}{|l|l|l|l|l|l|l|l|l|l|l|}
\hline
\multicolumn{1}{|c|}{\textit{\textbf{Category}}} & \textbf{1-NN} & \textbf{\begin{tabular}[c]{@{}l@{}}OC\\ SVM\end{tabular}} & \textbf{\begin{tabular}[c]{@{}l@{}}K\\ Means\end{tabular}} & \textbf{\begin{tabular}[c]{@{}l@{}}AE\\ MSE\end{tabular}} & \textbf{VAE} & \textbf{\begin{tabular}[c]{@{}l@{}}AE\\ SSIM\end{tabular}} & \textbf{\begin{tabular}[c]{@{}l@{}}Ano\\ GAN\end{tabular}} & \textbf{\begin{tabular}[c]{@{}l@{}}CNN\\ Feat.\\  Dic\end{tabular}} & \textbf{\begin{tabular}[c]{@{}l@{}}Uni.\\ Stud.\end{tabular}} & \textbf{\begin{tabular}[c]{@{}l@{}}VT-ADL\\ (Ours)\end{tabular}} \\ \hline
\textbf{Carpet} & 0.512 & 0.355 & 0.253 & 0.456 & 0.501 & 0.647 & 0.204 & 0.469 & 0.695 & \textbf{0.773} \\ \hline
\textbf{Grid} & 0.228 & 0.125 & 0.107 & 0.582 & 0.224 & 0.849 & 0.226 & 0.183 & 0.819 & \textbf{0.871} \\ \hline
\textbf{Leather} & 0.446 & 0.306 & 0.308 & 0.819 & 0.635 & 0.561 & 0.378 & 0.641 & \textbf{0.819} & 0.728 \\ \hline
\textbf{Tile} & 0.822 & 0.722 & 0.779 & 0.897 & 0.87 & 0.175 & 0.177 & 0.797 & \textbf{0.912} & 0.796 \\ \hline
\textbf{Wood} & 0.502 & 0.336 & 0.411 & 0.727 & 0.628 & 0.605 & 0.386 & 0.621 & 0.725 & \textbf{0.781} \\ \hline
\textbf{Bottle} & 0.898 & 0.85 & 0.495 & 0.91 & 0.897 & 0.834 & 0.62 & 0.742 & 0.918 & \textbf{0.949} \\ \hline
\textbf{Cable} & 0.806 & 0.431 & 0.513 & 0.825 & 0.654 & 0.478 & 0.383 & 0.558 & \textbf{0.865} & 0.776 \\ \hline
\textbf{Capsule} & 0.631 & 0.554 & 0.387 & 0.862 & 0.526 & 0.86 & 0.306 & 0.306 & \textbf{0.916} & 0.672 \\ \hline
\textbf{Hazelnut} & 0.861 & 0.616 & 0.698 & 0.917 & 0.878 & 0.916 & 0.698 & 0.844 & \textbf{0.937} & 0.897 \\ \hline
\textbf{Metal Nut} & 0.705 & 0.319 & 0.351 & 0.83 & 0.576 & 0.603 & 0.32 & 0.358 & \textbf{0.895} & 0.726 \\ \hline
\textbf{Pill} & 0.725 & 0.544 & 0.514 & 0.893 & 0.769 & 0.83 & 0.776 & 0.46 & \textbf{0.935} & 0.705 \\ \hline
\textbf{Screw} & 0.604 & 0.644 & 0.55 & 0.754 & 0.559 & 0.887 & 0.466 & 0.277 & \textbf{0.928} & \textbf{0.928} \\ \hline
\textbf{Toothbrush} & 0.675 & 0.538 & 0.337 & 0.822 & 0.693 & 0.784 & 0.749 & 0.151 & 0.863 & \textbf{0.901} \\ \hline
\textbf{Transistor} & 0.68 & 0.496 & 0.399 & 0.728 & 0.626 & 0.725 & 0.549 & 0.628 & 0.701 & \textbf{0.796} \\ \hline
\textbf{Zipper} & 0.512 & 0.355 & 0.253 & 0.839 & 0.549 & 0.665 & 0.467 & 0.703 & \textbf{0.933} & 0.808 \\ \hline
\textit{Means} & 0.64 & 0.479 & 0.423 & 0.79 & 0.639 & 0.694 & 0.443 & 0.515 & 0.857 & 0.807 \\ \hline
\end{tabular}
\caption{Comparative results on MVTec dataset. Comparative results taken from~\cite{bergmann2020uninformed}.}
\label{tab:mvtec-DR-conv-deco-results}
\end{table*}


\begin{table}[h]
\centering
\begin{tabular}{|l|l|l|l|l|}
\hline
\textbf{Prdt} & \textbf{\begin{tabular}[c]{@{}l@{}}PRO\\ Score\\ ours\end{tabular}} & \textbf{\begin{tabular}[c]{@{}l@{}}PR\\ AUC\\ ours\end{tabular}} & \begin{tabular}[c]{@{}l@{}}AE\\ MSE\end{tabular} & \multicolumn{1}{c|}{\begin{tabular}[c]{@{}c@{}}AE\\ MSE+SSIM\end{tabular}} \\ \hline
\textbf{0} & \textit{0.92} & 0.99 & 0.49 & 0.53 \\ \hline
\textbf{1} & \textit{0.89} & 0.94 & 0.92 & 0.96 \\ \hline
\textbf{2} & \textit{0.86} & 0.77 & 0.95 & 0.89 \\ \hline
\textit{Mean} & \textit{0.89} & \textbf{0.90} & 0.78 & 0.79 \\ \hline
\end{tabular}
\caption{Results on BTAD dataset.  We also compare our PR-AUC with the results of convolutional autoencoders trained with MSE loss and MSE+SSIM loss.}
\label{tab:result-BTAD}
\end{table}

Table \ref{tab:mvtec-DR-conv-deco-results} shows the results results for MVTec dataset. The value shows the PRO curve up to an average false positive rate per-pixel of 30\% is reported. It measures the average overlap of each ground truth region with the predicted anomaly region for multiple thresholds. Our proposed methods performed at par with the most recent state of the art algorithms (results taken from~\cite{bergmann2020uninformed}), and even outperformed them in 7 product categories.
for our newly published BTAD dataset, we are also reporting the first results in table \ref{tab:result-BTAD} with the similar model configuration as of MVTec. For comparison we also report PR-AUC of a basic convolutional autoencoder on BTAD with MSE and MSE+SSIM loss.

\subsection{Gaussian mixture model tuning}
\label{ablation}
\begin{figure}[ht]
	\centering
	\includegraphics[width=0.4\textwidth]{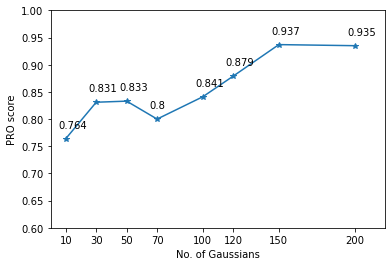}
	\caption{Plot shows the PRO score for the different no of Gaussians used in the Gaussian approximation.}
	\label{fig:no-of-gaus-ablation}
\end{figure}

Here we justify the choice of number of Gaussians for our mixture model. For this we trained on MVTec dataset with increasing number of Gaussians and calculated the PRO-score (fig.\ref{fig:no-of-gaus-ablation}). we found that with increasing number of Gaussians, PRO-score increases and then becomes constant. We also tried to see effect of noise addition in the transformer encoded features for generalisation. With noise added, PRO score with 150 Gaussians is \textit{0.897} in contrast to \textit{0.807} without noise. Hence, noise addition actually helps in generalizing the learning procedure.


\section{Conclusions}
We proposed a transformer-based framework which uses reconstruction and patch-based learning for image anomaly detection and localization. The anomalies can be detected at a global level using a reconstruction-based approach, and can be localized with the application of a Gaussian mixture model applied to the encoded image patches.
The achieved results are at par with or outperform other state-of-the-art techniques. We also published BTAD, a real world industrial dataset for the anomaly detection task. 

\bibliographystyle{IEEEtran}
\bibliography{mybibliography}

\end{document}